\documentclass[10pt,conference]{IEEEtran}
\IEEEoverridecommandlockouts
\usepackage{cite}
\usepackage{amsmath,amssymb,amsfonts}
\usepackage{algorithmic}
\usepackage{graphicx}
\usepackage{textcomp}
\usepackage{xcolor}
\usepackage[hyphens]{url}
\usepackage[hidelinks]{hyperref}
\usepackage{algorithm}
\usepackage{bm}
\usepackage{tikz}
\usetikzlibrary{positioning,arrows.meta,calc}

\usepackage{booktabs}
\usepackage{array}
\usepackage{subcaption}
\usepackage{float}
\newcommand{\R}{\mathbb{R}}
\newcommand{\clip}{\operatorname{clip}}
\newcommand{\rowmax}{\operatorname{rowmax}}

\newcommand{\MM}{\operatorname{MM}}

\newcommand{\EtwoMone}{\mathrm{E2M1}}

\newcommand{\Quant}{\operatorname{Quant}}

\title{EFQ-Softmax: Exp-Free Quantization for Softmax}
\hypersetup{
    pdftitle={EFQ-Softmax: Exp-Free Quantization for Softmax},
    pdfauthor={},
    pdfsubject={},
    pdfkeywords={}
}
\author{%
\IEEEauthorblockN{%
Haohui Han\textsuperscript{1},
Yuming Wan\textsuperscript{2},
Hongni Wang\textsuperscript{3},
Pengcheng Xie\textsuperscript{2}\\[0.25em]
Xiaodong Yan\textsuperscript{1,*},
Runqi You\textsuperscript{1},
Wencong Zhang\textsuperscript{1}}
\IEEEauthorblockA{%
\textsuperscript{1}Xi'an Jiaotong University\\
\textsuperscript{2}Huawei Technologies Co., Ltd\\
\textsuperscript{3}Shandong University of Finance and Economics\\[0.35em]
\small\texttt{hhhan200001@163.com}, \texttt{wanyuming3@huawei.com}, \texttt{wanghongnisd@126.com}\\
\small\texttt{xiepengcheng2@huawei.com}, \texttt{yanxiaodong@xjtu.edu.cn}\\
\small\texttt{18174543996@163.com}, \texttt{2161934877@qq.com}}
\thanks{\textsuperscript{*}Corresponding author: Xiaodong Yan (\href{mailto:yanxiaodong@xjtu.edu.cn}{yanxiaodong@xjtu.edu.cn}).}%
}
\hypersetup{pdfauthor={Haohui Han, Yuming Wan, Hongni Wang, Pengcheng Xie, Xiaodong Yan, Runqi You, Wencong Zhang}}

\date{}

\begin{document}
\maketitle
\thispagestyle{plain}
\pagestyle{plain}

\begin{abstract}

Low-bit attention accelerates Transformer inference by moving the $QK^\top$ and $PV$ matrix multiplications to FP8 or FP4 matrix engines. However, the softmax probability path often still evaluates shifted-score exponentials in higher precision, forms a temporary
probability block, and then quantizes it before the low-bit $PV$ multiplication. This exp-then-quantize path creates a mismatch between a high-precision probability producer and a low-bit matrix consumer.

This paper proposes \textbf{EFQ-Softmax} (\textbf{Exp-Free Quantization for Softmax}), a low-bit probability-generation method that directly maps shifted attention scores to block-scaled E2M1 operands. For each microscaling block, EFQ-Softmax selects an exponent-only scale from the local maximum, maps the shifted scores to a normalized residual domain, and generates nonnegative E2M1 probability codes using a single affine rule. The resulting operand is used consistently in both the $\widetilde{\bm{P}}\bm{V}$ numerator update and the $\widetilde{\bm{P}}\bm{1}$ denominator update. Meanwhile, the FlashAttention-style row-maximum update, historical rescaling, high-precision accumulation, and final normalization remain unchanged.

We evaluate the end-to-end quality of EFQ-Softmax across language, vision-language, and text-to-video workloads, and separately evaluate its kernel-level performance on the \mbox{Ascend~950PR} vector unit. Specifically, we evaluate EFQ-Softmax on Qwen3-8B, Qwen3-VL-8B-Instruct, and WAN2.2-TI2V-5B. EFQ-Softmax improves the Qwen3-8B seven-task mean from $0.6749$ with MXFP4 to $0.6773$ and the Qwen3-VL nine-task mean from $0.7826$ to $0.8000$. On WAN2.2, it maintains temporal consistency and visual quality comparable to the FP16 and MXFP4 baselines under VBench. On the \mbox{Ascend~950PR} vector unit, EFQ-Softmax reduces the vector-stage latency of the fused probability-generation kernel by \textbf{$40.33\%$} on average across sequence lengths from $16\mathrm{K}$ to $128\mathrm{K}$. These results show that direct low-bit probability generation can replace the conventional exp-then-quantize path while preserving end-to-end model quality.
\end{abstract}

\section{Introduction}

Self-attention is a central operator in modern Transformer
models~\cite{vaswani2017attention}. In a standard implementation, its arithmetic cost grows quadratically with the sequence length, and the score and probability matrices may also require quadratic storage. These costs make attention expensive in long-context language models,
vision-language models, and video-generation models. FlashAttention reduces memory traffic by computing attention in tiles and maintaining the softmax statistics through an online recurrence, without materializing the full attention matrix~\cite{dao2022flashattention}.
FlashAttention-2 further improves parallelism and work
partitioning~\cite{dao2024flashattention}, while FlashAttention-3 uses asynchronous execution and low-precision matrix multiplication on recent hardware~\cite{shah2024flashattention}. In parallel,
low-precision formats such as FP8 and microscaling FP4 provide compact operands for modern matrix engines
~\cite{micikevicius2022fp8,rouhani2023microscaling}. Quantized attention systems such as SageAttention and SageAttention3 show that the two matrix multiplications in attention can be executed at low precision while maintaining model quality across language, image, and video
workloads~\cite{zhang2024sageattention,zhang2026sageattention3}. As these matrix multiplications become faster, the probability-generation operations between score computation and value multiplication account for a larger part of the attention execution time.

In a FlashAttention-style kernel, each score tile is first shifted by an updated row maximum for numerical stability. Elementwise exponentials are then evaluated to generate the current unnormalized probability weights. In a conventional FP4 attention path, these weights are first produced in FP16 or FP32 and are then quantized into
a block-scaled E2M1 representation for the low-bit value
multiplication. This exp-then-quantize path therefore contains three separate steps: evaluating dense elementwise exponentials, forming a temporary high-precision probability tile, and quantizing the tile into
a low-bit operand. The resulting pipeline uses a high-precision probability producer even though the following matrix multiplication consumes low-precision values. Existing FP4 attention methods improve the scaling and quantization of probability values, but still compute
the exponential values before converting them into a low-bit representation~\cite{zhang2026sageattention3}. This producer-consumer mismatch motivates bypassing the temporary high-precision probability representation and directly generating the low-bit operand from the shifted scores. However, doing so is not merely a scalar exponential-approximation problem. The generated values must be directly usable by the low-bit value multiplication, and the same approximation must be used in both the softmax numerator and denominator. At the same time, the online
recurrence must preserve its row-maximum update, historical rescaling, high-precision accumulation, and final normalization.

\begin{figure}[!t]
\centering
\makebox[\columnwidth][l]{%
\hspace*{-0.04\columnwidth}%
\resizebox{1.04\columnwidth}{!}{%
\begin{tikzpicture}[
    >=Stealth,
    font=\fontsize{12pt}{14.8pt}\selectfont,
    node distance=1.25em and 1.35em,
    box/.style={
        rectangle,
        draw,
        rounded corners=2pt,
        inner xsep=0.80em,
        inner ysep=0.58em,
        align=center
    },
    proc/.style={
        rectangle,
        draw,
        inner xsep=0.75em,
        inner ysep=0.52em,
        align=center
    },
    branch/.style={
        rectangle,
        draw,
        inner xsep=0.70em,
        inner ysep=0.52em,
        align=center
    },
    plain/.style={
        inner sep=0.20em,
        align=center
    }
]

\node (score) [box]
{
Current score tile
\\[0.18em]
$\displaystyle
\bm{S}_i^j
=
\frac{\bm{Q}_i\bm{K}_j^{\top}}{\sqrt{d}}
+
\bm{M}_i^j
$
};

\node (rowmax) [
    proc,
    below=of score
]
{
Row-maximum update
\\[0.18em]
$\displaystyle
\bm{m}_i^j
=
\max
\left\{
\bm{m}_i^{j-1},
\operatorname{rowmax}(\bm{S}_i^j)
\right\}
$
};

\draw[->] (score) -- (rowmax);

\node (shift) [
    proc,
    below=of rowmax
]
{
Shifted score tile
\\[0.18em]
$\bm{x}_i^j=\bm{S}_i^j-\bm{m}_i^j$
};

\draw[->] (rowmax) -- (shift);

\node (EFQ-Softmax) [
    box,
    below=of shift,
    text width=25em
]
{
\textbf{EFQ-Softmax Probability Generation}
\\[0.10em]
Replaces elementwise exp and post-exp quantization
\\[0.32em]
Block maximum and exponent-only scale
\\
Residual log-domain normalization
\\
Affine E2M1 code generation
\\[0.32em]
\textbf{Block-scaled E2M1 operand}
\\[-0.02em]
$\displaystyle
\left(
\bm{Q}_{P,i}^{j,\mathrm{E2M1}},
\bm{s}_{P,i}^{j}
\right)
$
};

\draw[->] (shift) -- (EFQ-Softmax);

\coordinate (highlightNW) at
    ($(EFQ-Softmax.north west)+(-0.35em,0.40em)$);

\coordinate (highlightSE) at
    ($(EFQ-Softmax.south east)+(0.35em,-0.35em)$);

\draw[
    red,
    densely dashed,
    rounded corners=10pt,
    line width=2.2pt
]
(highlightNW) rectangle (highlightSE);

\node[
    red,
    anchor=south west,
    font=\fontsize{13pt}{15.6pt}\selectfont
] at ($(EFQ-Softmax.north west)+(-0.10em,0.52em)$)
{Proposed path};

\coordinate (fork) at
    ($(EFQ-Softmax.south)+(0,-2.0em)$);

\node (numerator) [
    branch,
    below=4.50em of EFQ-Softmax.south,
    xshift=-9.3em
]
{
Numerator update
\\[0.18em]
$\displaystyle
\Delta\bm{A}_i^j
=
\widetilde{\bm{P}}_i^j\bm{V}_j
$
};

\node (denominator) [
    branch,
    below=4.50em of EFQ-Softmax.south,
    xshift=9.3em
]
{
Denominator update
\\[0.18em]
$\displaystyle
\Delta\bm{l}_i^j
=
\widetilde{\bm{P}}_i^j\bm{1}
$
};

\draw[->]
    (EFQ-Softmax.south)
    -- (fork)
    -| (numerator.north);

\draw[->]
    (EFQ-Softmax.south)
    -- (fork)
    -| (denominator.north);

\node (value) [
    plain,
    right=0.45em of numerator,
    text width=7.2em
]
{
Existing low-bit
\\
value operand
\\[-0.02em]
$\bm{V}_j$
};

\draw[->]
    (value.west)
    --
    (numerator.east);

\coordinate (branchmid) at
    ($(numerator.south)!0.5!(denominator.south)$);

\node (update) [
    box,
    below=1.35em of branchmid,
    text width=20.5em
]
{
\textbf{Preserved online recurrence}
\\[0.25em]
Row-level historical rescaling
\\
FP16/FP32 numerator and
\\
denominator accumulation
};

\draw[->] (numerator) -- (update);
\draw[->] (denominator) -- (update);

\node (output) [
    box,
    below=of update
]
{
Final normalization
\\[0.18em]
$\displaystyle
\widetilde{\bm{O}}_i
=
\frac{\widetilde{\bm{A}}_i}
     {\widetilde{\bm{l}}_i}
$
};

\draw[->] (update) -- (output);

\end{tikzpicture}%
}%
}

\caption{
EFQ-Softmax directly generates a shared
block-scaled E2M1 operand within online attention.
}
\label{fig:EFQ-Softmax-online-flow}
\end{figure}
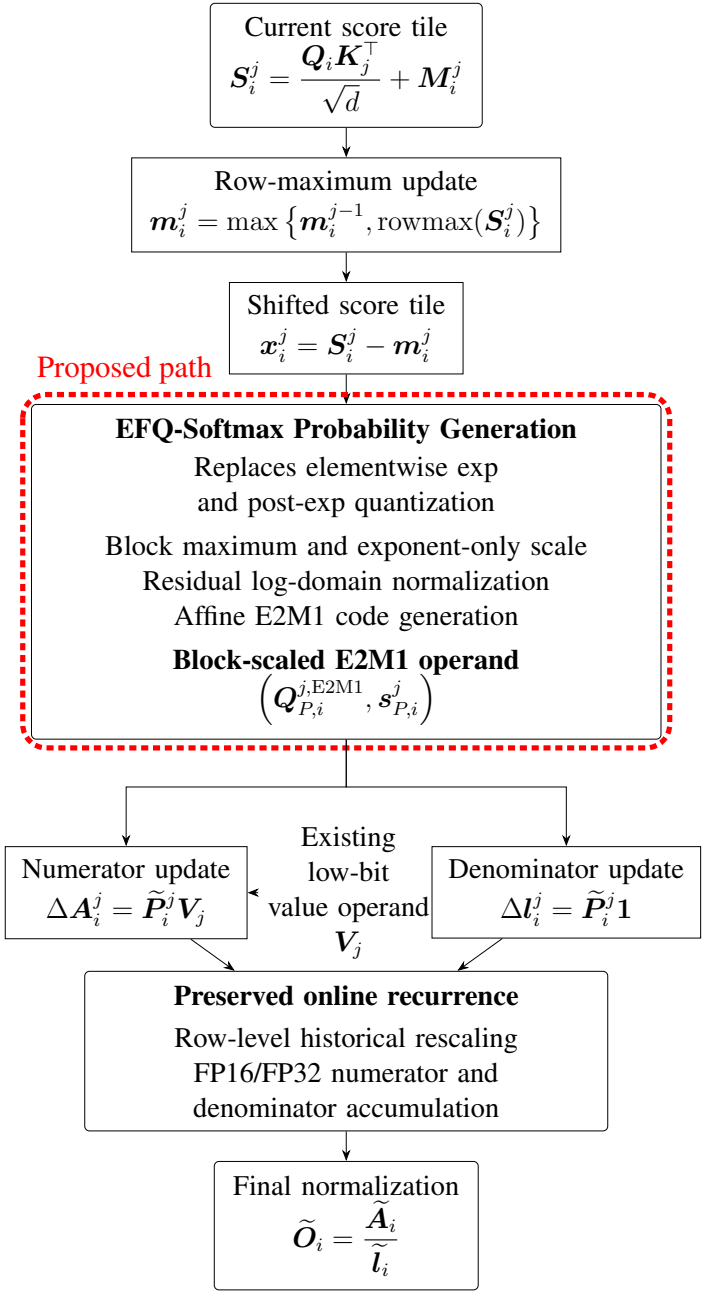

\begin{figure*}[!t]
    \centering
    \includegraphics[
    width=\textwidth,
    trim=4mm 4mm 4mm 4mm,
    clip
]
    {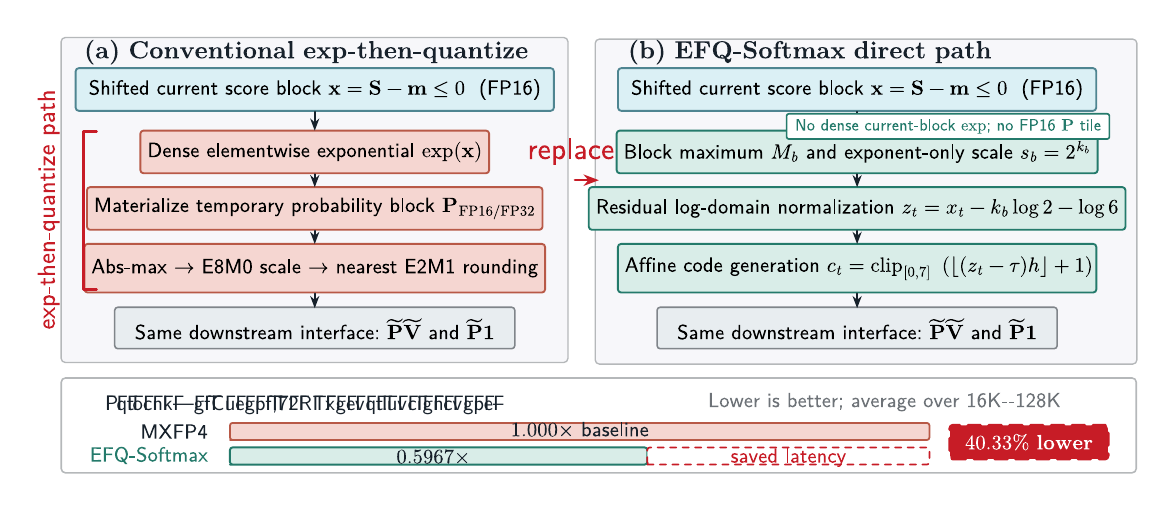}
    \caption{Conventional exp-then-quantize versus EFQ-Softmax. EFQ-Softmax directly generates block-scaled E2M1 probabilities from shifted scores, reducing \mbox{Ascend~950PR} vector-stage latency by 40.33\%.}
    \label{fig:path_comparison}
\end{figure*}

To address this problem, we propose  EFQ-Softmax, a low-bit
probability-generation method that directly converts shifted attention scores into a block-scaled E2M1 operand. Here, ``Exp-Free'' refers to removing the dense elementwise exponential computation for the current
score block, while retaining the row-level exponential required for historical rescaling. For each microscaling block, EFQ-Softmax selects an exponent-only scale from the local maximum, maps the shifted scores to a scale-normalized residual domain, and generates nonnegative E2M1
codes using a single affine index rule. Together with the block-level E8M0 scale, these codes form the final MXFP4 probability operand and are used consistently in both the numerator and denominator updates. This design eliminates the intermediate FP16 or FP32 probability tile and the separate post-exponential quantization step, while preserving the row-maximum update, historical rescaling, high-precision accumulation, and final normalization of online attention.

Fig.~\ref{fig:EFQ-Softmax-online-flow} illustrates how EFQ-Softmax is integrated into the online attention recurrence, while Fig.~\ref{fig:path_comparison} compares its direct score-to-E2M1 generation path with the conventional exp-then-quantize path.

We evaluate EFQ-Softmax on Qwen3-8B~\cite{yang2025qwen3}, Qwen3-VL-8B-Instruct~\cite{bai2025qwen3}, and WAN2.2-TI2V-5B~\cite{wan2025wan}. On Qwen3-8B, EFQ-Softmax improves the
seven-task mean score from $0.6749$ with MXFP4 to $0.6773$. On Qwen3-VL-8B-Instruct, it improves the nine-task mean score from $0.7826$ to $0.8000$ and achieves better results on all six referring-expression grounding splits. On WAN2.2, EFQ-Softmax maintains temporal consistency and visual quality comparable to the
FP16 and MXFP4 baselines under VBench~\cite{huang2024vbench}. We also evaluate the probability-generation path on the \mbox{Ascend~950PR} vector unit.
Across sequence lengths from 16K to 128K, EFQ-Softmax reduces the vector-stage latency of the fused probability-quantization kernel by $40.33\%$ on average. These results show that EFQ-Softmax shortens the probability-generation path while preserving numerical accuracy and end-to-end model quality.

Our main contributions are summarized as follows:
\begin{itemize}

    \item We propose EFQ-Softmax, a low-bit probability-generation method that fuses exponential approximation and MXFP4 probability quantization. It maps shifted attention scores to E2M1 probability codes through a single affine transformation and rounding, together with a block-level E8M0 scale.

    \item Through integrating EFQ-Softmax into the FlashAttention-style online attention computation, the method preserves row-maximum and row-sum state updates, avoids storing full $S$ or $P$ matrices, and directly feeds the generated MXFP4 probability block into the low-bit $PV$ multiplication.

    \item  EFQ-Softmax is evaluated on both controlled simulations and real model workloads. Experiments on Qwen3-8B, Qwen3-VL, and Wan2.2 show that the proposed method improve attention throughput while preserving numerical accuracy and end-to-end output quality.
\end{itemize}

The rest of this paper is organized as follows. Section~\ref{sec:methodology} presents the design of EFQ-Softmax, including the block-scaled E2M1 representation, exponent-only scale selection, residual-domain normalization, affine code generation, and the EFQ-Softmax-16 variant. Section~\ref{sec:integration} describes the integration of EFQ-Softmax into FlashAttention-style online attention, including the numerator and denominator updates, algorithm summary, parameter calibration, and implementation boundary. Section~\ref{sec:experiments} evaluates EFQ-Softmax on language, vision-language, text-to-video, and kernel-level workloads. Section~\ref{sec:conclusion} concludes the paper.

\section{EFQ-Softmax Methodology}
\label{sec:methodology}

\subsection{Design Scope and Overview}

\paragraph{Targeted probability path.}
For one attention head, let $Q\in\R^{N_q\times d}$, $K\in\R^{N_k\times d}$, and $V\in\R^{N_k\times d_v}$. The exact attention output is
\[
O=\operatorname{softmax}\left(\frac{QK^\top}{\sqrt d}\right)V .
\]
Let $S=QK^\top/\sqrt d$ denote the score matrix. Stable softmax evaluates each row after subtracting the row maximum:
\[
P_{it}=\frac{\exp(S_{it}-m_i)}{\sum_r\exp(S_{ir}-m_i)},\qquad
m_i=\max_r S_{ir} .
\]
The elementwise term that must be generated before normalization is therefore
\begin{equation}
p_{it}=\exp(x_{it}),\qquad x_{it}=S_{it}-m_i\leq 0 .
\label{eq:shifted-x}
\end{equation}
We refer to $x_{it}=S_{it}-m_i$ as a row-max-shifted
attention score, or simply a shifted score.

EFQ-Softmax modifies only the probability-generation path. It replaces the elementwise computation of $\exp(x)$ for the current score block and the subsequent probability quantization with direct generation of a low-bit probability operand. The attention formulation, value-side representation, and precision of the online accumulators remain unchanged from the surrounding attention kernel. The value block is not modified by EFQ-Softmax; it only consumes the generated probability operand in the subsequent (PV) multiplication.

In a low-bit attention path, the $PV$ consumer often expects a low-bit probability operand. A conventional construction first evaluates Eq.~\eqref{eq:shifted-x} in a higher-precision format and then quantizes the resulting probability block:
\begin{equation}
\label{eq:post-softmax-path}
x\rightarrow \exp(x)\rightarrow P_{\mathrm{FP16/FP32}}
\rightarrow \Quant_{\EtwoMone}(P)\rightarrow Q_P^{\EtwoMone}.
\end{equation}
EFQ-Softmax replaces the exp-then-quantize sequence by direct probability-code generation:
\begin{equation}
\label{eq:d8-short-path}
x\rightarrow c(x)\rightarrow Q_P^{\EtwoMone} .
\end{equation}
The generated operand is then shared by the numerator branch $\widetilde P V$ and the denominator branch $\widetilde P\mathbf 1$. The outputs of these two branches are accumulated in FP16 or FP32, as in standard low-bit matrix-multiply pipelines.

\paragraph{Online attention recurrence.}

To describe how EFQ-Softmax operates in a tiled online-attention kernel, we now move from the elementwise notation above to a blockwise formulation. FlashAttention processes the score matrix by query blocks and key/value blocks. For query block $i$ and key/value block $j$, the score block is
\[
S_i^j=\frac{Q_iK_j^\top}{\sqrt d}+M_i^j,
\]
where $M_i^j$ denotes an optional mask term. The online state consists of a row maximum $m_i^j$, a numerator accumulator $A_i^j$, and a denominator accumulator $l_i^j$. The row maximum update is
\begin{equation}
\label{eq:rowmax-update}
m_i^j=\max\left\{m_i^{j-1},\rowmax(S_i^j)\right\} .
\end{equation}
Using $x_i^j=S_i^j-m_i^j$, the exact current unnormalized probability block is
\[
P_i^j=\exp(x_i^j) .
\]
The historical state must be rescaled when the row maximum changes. With
\begin{equation}
\label{eq:alpha}
\alpha_i^j=\exp(m_i^{j-1}-m_i^j),
\end{equation}
the exact online update is
\begin{align}
A_i^j&=\alpha_i^jA_i^{j-1}+P_i^jV_j,\label{eq:exact-A}\\
l_i^j&=\alpha_i^jl_i^{j-1}+P_i^j\mathbf 1 .\label{eq:exact-l}
\end{align}
EFQ-Softmax keeps Eqs.~\eqref{eq:rowmax-update}-\eqref{eq:exact-l}. Only the current-block probability block $P_i^j=\exp(x_i^j)$ is replaced by a block-scaled low-bit approximation.

\paragraph{Required probability operand.}

These constraints define the required EFQ-Softmax output. The generated
operand must be directly consumed by the low-bit \(PV\) path, reused by
the denominator update, and accumulated into the existing high-precision
online state. We therefore represent the current unnormalized probability
block as
\[
\widetilde P_i^j\approx s_{P,i}^j Q_{P,i}^{j,\EtwoMone} .
\]
Here $Q_{P,i}^{j,\EtwoMone}$ is a nonnegative E2M1 code block and $s_{P,i}^j$ is a block scale. The rest of this section describes how
$Q_{P,i}^{j,\EtwoMone}$ and $s_{P,i}^j$ are generated directly
from the shifted score block $x_i^j=S_i^j-m_i^j$. 

\subsection{Block-Scaled E2M1 Probability Representation}

\paragraph{Nonnegative E2M1 code set.}

The E2M1 format contains a small number of representable values. Since attention probabilities are nonnegative, EFQ-Softmax only uses the nonnegative code set
\[
\mathcal C_8=
\left\{0,\frac12,1,\frac32,2,3,4,6\right\} .
\label{eq:c8}
\]

The index set is \(\{0,\ldots,7\}\), and \(\mathcal C_8[c]\) denotes the
numerical value represented by code \(c\). This notation is only used to
state the mathematical value of each bit pattern. In the data path, the
generated code is packed as a 4-bit E2M1 probability operand rather than
used as an index for a runtime value lookup.

For each microscaling block $b$, EFQ-Softmax writes
$
\widetilde p_t=s_bq_t,
q_t\in\mathcal C_8,
t\in b .
$
The scale $s_b$ determines the exponent range of the block, while the code value $q_t$ determines the local mantissa-like value. In the probability path, this is sufficient because all inputs satisfy $x_t\leq 0$ after the row-maximum shift and the largest exact value inside a row is at most one.

\paragraph{Exponent-Only Scale Rule}

Let
$\displaystyle
M_b=\max_{t\in b}x_t
$
be the largest shifted score inside a microscaling block. Since the largest value in $\mathcal C_8$ is $6$, the scale is chosen so that $6s_b$ is close to $e^{M_b}$. EFQ-Softmax restricts the scale to an exponent-only value
$s_b=2^{k_b} .$
The scale exponent is selected by
\begin{equation}
\label{eq:k-rule}
k_b=\left\lfloor\frac{M_b+\log(2/9)}{\log 2}\right\rfloor .
\end{equation}
Equivalently,
\begin{equation}
2^{k_b}\leq \frac{2e^{M_b}}{9}<2^{k_b+1} .
\label{eq:k-interval}
\end{equation}
The constant $\log(2/9)$ is part of the scale rule. It is not a fitted EFQ-Softmax parameter. Its role is to choose a power-of-two scale around the largest local probability value under the E2M1 maximum code $6$. The only fitted EFQ-Softmax parameters are the affine
code-generation parameters \((\tau,h)\), introduced below.

This exponent-only scale is compatible with the probability-side data path. It avoids a general floating-point scale generation step and provides a compact representation that can be consumed together with the generated E2M1 code. The scale is shared by all elements in the microscaling block; therefore the elementwise decision that remains after scaling is only an index-generation problem.

\paragraph{Residual Log-Domain Normalization}

After the scale has been selected, each shifted score is converted into a residual log-domain value
\begin{equation}
\label{eq:z-def}
z_t=x_t-\log(6s_b)=x_t-k_b\log2-\log6 .
\end{equation}
This residual satisfies
\[
e^{x_t}=s_b\cdot 6e^{z_t} .
\label{eq:residual-exp}
\]
Thus the remaining operation is to approximate $6e^{z_t}$ by one of the values in $\mathcal C_8$. The residual $z_t$ is the proper input to the code-generation rule because it has already absorbed the block scale. Using $x_t$ directly without this normalization would mix two tasks: block-level dynamic-range selection and element-level code selection.

The largest residual in block $b$ is
\[
z_{\max,b}=M_b-k_b\log2-\log6,
\]
From Eq.~\eqref{eq:k-interval}, $z_{\max,b}$ remains in a bounded interval determined by the power-of-two rounding of the scale. This bounded residual range is the reason that a single pair of EFQ-Softmax parameters can be reused across many blocks. The exact shifted scores may have a large dynamic range across rows and layers, but the residual code-generation input is locally normalized.

\paragraph{Microscaling blocks.}

The microscaling block size is a design parameter of the probability path. For a flattened local score block, the partition can be written as
\[
\mathcal B=\{b_1,b_2,\ldots,b_R\},
\qquad |b_r|=B .
\]
For each $b_r$, the scale exponent $k_{b_r}$ is computed once and then reused by all elements in the block. A smaller $B$ gives more local scaling and can reduce approximation error, while a larger $B$ reduces scale metadata and scale-generation work. EFQ-Softmax does not rely on a particular value of $B$ in the mathematical definition, and the implementation keeps $B$ consistent with the low-bit matrix operand layout.

\subsection{EFQ-Softmax Probability Code Generation}
\label{sec:EFQ-Softmax-codegen}
Given the residual \(z_t\) from Eq.~\eqref{eq:z-def}, EFQ-Softmax generates
a 4-bit E2M1 probability code with a single affine rule:
\begin{equation}
\label{eq:d8-index}
c_t=\clip_{[0,7]}\left(\left\lfloor(z_t-\tau)h\right\rfloor+1\right) .
\end{equation}
The generated $c_t$ is the probability code. Operationally, it is stored as a 4-bit pattern and interpreted as a nonnegative E2M1 operand:
\[
Q_{P,t}^{\EtwoMone}\leftarrow \operatorname{reinterpret}_{\EtwoMone}(c_t) .
\]
Mathematically, the represented probability value is
$
\widetilde p_t=s_b\mathcal C_8[c_t].
$
Here \(\mathcal C_8[c_t]\) denotes the numerical value of the E2M1 bit
pattern and does not imply a runtime lookup of a high-precision
exponential value.

\begin{figure*}[!t] \centering \includegraphics[
    width=\textwidth,
    trim=3mm 2mm 3mm 2mm,
    clip
]{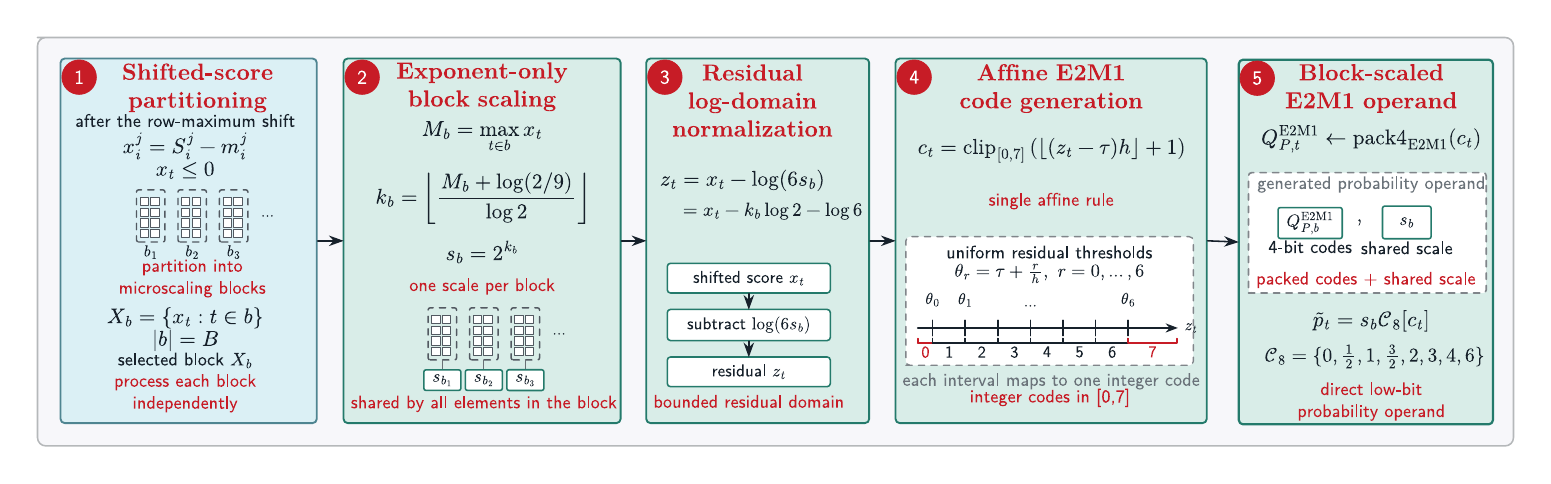}
\caption{ Overview of the five-stage EFQ-Softmax probability-generation procedure.  } \label{fig:efq-step-by-step} \end{figure*}

Combining the scale rule in Eq.~\eqref{eq:k-rule}, the residual definition
in Eq.~\eqref{eq:z-def}, and the index rule in Eq.~\eqref{eq:d8-index},
EFQ-Softmax implements the direct path
\[
x_t \rightarrow z_t \rightarrow c_t \rightarrow Q_{P,t}^{\EtwoMone}.
\]
Equivalently, for \(t\in b\), the represented probability value is
\[
\begin{aligned}
\widetilde p_t
&=2^{k_b}\mathcal C_8[c_t],\\
c_t
&=\clip_{[0,7]}\left(\left\lfloor
\left(x_t-k_b\log 2-\log 6-\tau\right)h
\right\rfloor+1\right).
\end{aligned}
\]
The generated result is already the low-bit probability operand consumed
by the following attention update, so EFQ-Softmax does not materialize an
intermediate FP16/FP32 probability block required in the intended path.

The parameters \(\tau\) and \(h\) define uniformly spaced thresholds in
the residual log domain. Ignoring clipping, code \(c\) is selected when
\[
\tau+\frac{c-1}{h}\leq z_t<\tau+\frac{c}{h}.
\]
The parameter \(\tau\) shifts the thresholds, while \(h\) controls their
spacing. Thus, EFQ-Softmax uses uniformly spaced residual thresholds together
with a fixed nonuniform E2M1 output set.
EFQ-Softmax does not use these thresholds to fetch a high-precision
exponential value. The output of Eq.~\eqref{eq:d8-index} is already the
final 4-bit E2M1 code consumed by the attention data path.

The clipping in Eq.~\eqref{eq:d8-index} bounds the generated code to the
nonnegative E2M1 range. Residuals below the lowest threshold map to code
\(0\), while residuals above the highest threshold map to code \(7\),
which represents the largest nonnegative E2M1 value \(6\).
The zero code is not a separate pruning rule; it is one representable
outcome of the block-scaled E2M1 probability operand. This saturation is
part of bounded index generation rather than an additional algorithmic
branch.

If \(6e^z\) were first computed and then nearest-rounded to
\(\mathcal C_8\), the resulting log-domain boundaries would be nonuniform
because the E2M1 values are nonuniform. EFQ-Softmax deliberately gives up this
scalar nearest-rounding rule and uses affine residual thresholds to
simplify code generation.

This design trades scalar nearest-rounding optimality for a simpler
probability-code generation path.  Nearest-code rounding after an explicit exponential requires computing $e^x$ and then deciding the final E2M1 value. EFQ-Softmax removes the explicit exponential and learns a direct residual threshold rule. The correctness target is not exact scalar exponential reconstruction. The relevant target is output-level attention accuracy after the numerator and denominator updates.

The parameter \(\tau\) controls the global threshold shift, and \(h\)
controls the threshold spacing. Their values are selected offline by the
calibration procedure described in Sec.~\ref{sec:parameter-calibration}.

The complete EFQ-Softmax probability-generation procedure is summarized in Figure~\ref{fig:efq-step-by-step}. The figure shows how shifted attention scores are transformed into packed block-scaled E2M1 probability operands through five consecutive stages. 

\subsection{EFQ-Softmax-LUT Variant}
\label{sec:EFQ-Softmax-table}
EFQ-Softmax-LUT uses the same block scale and the same nonnegative E2M1
output set as EFQ-Softmax, but introduces a 16-level intermediate index before
folding to the final 8-code E2M1 output.
The intermediate index is
\[
j_{16}(z_t)=
\clip_{[0,15]}\left(\left\lfloor(z_t-\tau_{16})h_{16}\right\rfloor+1\right),
\]where
$\tau_{16}=\log(1/24),\quad h_{16}=\frac{14}{\log24}.
$
The final E2M1 code is obtained by the monotone folding table
\[
C_{16}=[0,1,1,1,1,1,2,2,3,3,4,5,5,6,7,7].
\]
Then the final code is
\[
c_t^{\mathrm{tab}}=C_{16}[j_{16}(z_t)],\qquad
Q_{P,t}^{\EtwoMone}\leftarrow \operatorname{pack4}_{\EtwoMone}(c_t^{\mathrm{tab}}).
\]
Mathematically, the represented probability value is
\(\widetilde p_t^{\mathrm{tab}}=s_b\mathcal C_8[c_t^{\mathrm{tab}}]\).
In contrast to EFQ-Softmax, this variant contains an explicit 16-to-8
remapping stage.

\begin{figure}[t]
    \centering
    \includegraphics[
        width=\columnwidth,
        trim=4mm 2mm 4mm 2mm,
        clip
    ]{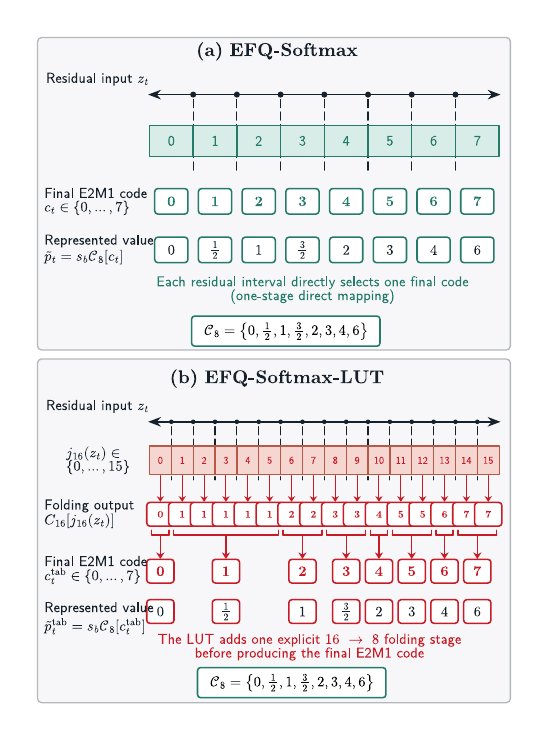}
    \caption{
        Affine code mapping in EFQ-Softmax and two-stage mapping
        in EFQ-Softmax-LUT.
    }
    \label{fig:efq-lut-mapping}
\end{figure}

Fig.~\ref{fig:efq-lut-mapping} compares the affine code mapping
of EFQ-Softmax with the two-stage mapping of EFQ-Softmax-LUT.

EFQ-Softmax-LUT is useful as an accuracy-oriented comparison point. It has finer intermediate boundary placement than the 8-code EFQ-Softmax rule, but it also introduces an additional remapping stage. For this reason, the main method remains EFQ-Softmax. EFQ-Softmax-LUT acts as an ablation that measures how much accuracy can be gained by increasing mapping flexibility at the probability-code generation point.

\section{Online Integration and Implementation}
\label{sec:integration}

\subsection{Integration with Online Attention}
\label{sec:online-integration}

\paragraph{Probability operand and consumers.}

For a score block $(i,j)$, EFQ-Softmax produces the probability operand
\[
\widetilde P_i^j\approx s_{P,i}^jQ_{P,i}^{j,\EtwoMone} .
\]
The numerator contribution is
\begin{equation}
\label{eq:numerator-basic}
\Delta A_i^j=\widetilde P_i^jV_j.
\end{equation}
If the value block is already represented by an existing low-bit path, $V_j\approx s_V^jQ_V^j$, the consumer can be written as
\begin{equation}
\label{eq:numerator-lowbit}
\Delta A_i^j\approx s_{P,i}^js_V^j\MM\left(Q_{P,i}^{j,\EtwoMone},Q_V^j\right).
\end{equation}
Equation~\eqref{eq:numerator-lowbit} only specifies the consumer of the generated probability operand. It is not a new value quantization rule. The contribution of EFQ-Softmax is the generation of $Q_{P,i}^{j,\EtwoMone}$ from the shifted score block.

The same probability operand is also used by the denominator branch:
\begin{equation}
\label{eq:denominator-basic}
\Delta l_i^j=\widetilde P_i^j\mathbf 1
\approx s_{P,i}^j\left(Q_{P,i}^{j,\EtwoMone}\mathbf 1\right).
\end{equation}
The denominator increment is not stored as an E2M1 value; the low-bit
operand provides the input to the row-sum operation, and the row-sum output
is accumulated in FP16 or FP32.

Using the same probability operand in Eqs.~\eqref{eq:numerator-basic} and \eqref{eq:denominator-basic} is important. If the numerator used EFQ-Softmax but the denominator used a separately reconstructed probability block, the online update would combine two different probability approximations. EFQ-Softmax instead uses one block-scaled probability operand for both branches.

\paragraph{Online recurrence.}
With the rescaling factor $\alpha_i^j$ from Eq.~\eqref{eq:alpha}, the EFQ-Softmax online update is
\begin{align*}
\widetilde A_i^j
&=\alpha_i^j\widetilde A_i^{j-1}+\Delta A_i^j,\\
\widetilde l_i^j
&=\alpha_i^j\widetilde l_i^{j-1}+\Delta l_i^j.
\end{align*}
The output after all key/value blocks are processed is
\[
\widetilde O_i=\frac{\widetilde A_i}{\widetilde l_i}.
\]
Thus the online structure of FlashAttention is preserved. EFQ-Softmax replaces only the current-block generation of $P_i^j$. The row maximum update, historical rescaling, high-precision accumulation, and final division remain the same.

\paragraph{Current-block exponential path.}

The current probability block normally requires elementwise exponentials for all entries of $x_i^j$. EFQ-Softmax removes these current-block elementwise exponentials and replaces them with affine index generation. The historical rescaling factor $\alpha_i^j=\exp(m_i^{j-1}-m_i^j)$ is a row-level term and remains part of the online recurrence. This distinction is important: the method does not remove every exponential operation in FlashAttention. It removes the dense elementwise exponential path used to materialize the current unnormalized probability block.

\subsection{Algorithm Summary}
\label{sec:algorithm-summary}

Algorithms~\ref{alg:d8-codegen} and~\ref{alg:d8-online} summarize the
online execution of EFQ-Softmax. Algorithm~\ref{alg:d8-codegen} describes
the probability-code generation rule for one microscaling block, while
Algorithm~\ref{alg:d8-online} shows how the generated operand is consumed
inside one online attention update. Algorithm~\ref{alg:param-search}
summarizes the offline calibration procedure used to select the global
parameters $(\tau,h)$ by minimizing the attention-output error over
representative calibration blocks.

\begin{algorithm}[t]
\caption{EFQ-Softmax Probability-Code Generation for One Microscaling Block}
\label{alg:d8-codegen}
\begin{algorithmic}[1]
\REQUIRE Shifted scores $\{x_t:t\in b\}$, EFQ-Softmax parameters $(\tau,h)$
\ENSURE Block scale $s_b$ and nonnegative E2M1 probability codes $\{Q_{P,t}^{\EtwoMone}:t\in b\}$
\STATE $M_b\leftarrow \max_{t\in b}x_t$
\STATE $k_b\leftarrow\left\lfloor(M_b+\log(2/9))/\log2\right\rfloor$
\STATE $s_b\leftarrow 2^{k_b}$
\FOR{each element $t\in b$}
    \STATE $z_t\leftarrow x_t-k_b\log2-\log6$
    \STATE $c_t\leftarrow\clip_{[0,7]}\left(\left\lfloor(z_t-\tau)h\right\rfloor+1\right)$
    \STATE $Q_{P,t}^{\EtwoMone}\leftarrow\operatorname{pack4}_{\EtwoMone}(c_t)$
\ENDFOR
\RETURN $s_b$, $\{Q_{P,t}^{\EtwoMone}:t\in b\}$
\end{algorithmic}
\end{algorithm}

\begin{algorithm}[t]
\caption{EFQ-Softmax Online Update for One Key/Value Block}
\label{alg:d8-online}
\begin{algorithmic}[1]
\REQUIRE $Q_i$, $K_j$, $V_j$, previous state $(m_i^{j-1},\widetilde A_i^{j-1},\widetilde l_i^{j-1})$
\ENSURE Updated state $(m_i^j,\widetilde A_i^j,\widetilde l_i^j)$
\STATE $S_i^j\leftarrow Q_iK_j^\top/\sqrt d$
\STATE $m_i^j\leftarrow \max\{m_i^{j-1},\rowmax(S_i^j)\}$
\STATE $x_i^j\leftarrow S_i^j-m_i^j$
\STATE Partition $x_i^j$ into microscaling blocks
\FOR{each microscaling block $b$}
    \STATE Generate $s_b$ and $Q_{P,b}^{\EtwoMone}$ by Algorithm~\ref{alg:d8-codegen}
\ENDFOR
\STATE Assemble $\widetilde P_i^j\approx s_{P,i}^jQ_{P,i}^{j,\EtwoMone}$ from all blocks
\STATE $\Delta A_i^j\leftarrow \widetilde P_i^jV_j$
\STATE $\Delta l_i^j\leftarrow \widetilde P_i^j\mathbf 1$
\STATE $\alpha_i^j\leftarrow\exp(m_i^{j-1}-m_i^j)$
\STATE $\widetilde A_i^j\leftarrow \alpha_i^j\widetilde A_i^{j-1}+\Delta A_i^j$
\STATE $\widetilde l_i^j\leftarrow \alpha_i^j\widetilde l_i^{j-1}+\Delta l_i^j$
\RETURN $m_i^j$, $\widetilde A_i^j$, $\widetilde l_i^j$
\end{algorithmic}
\end{algorithm}

\begin{algorithm}[t]
\caption{Offline Selection of EFQ-Softmax Parameters}
\label{alg:param-search}
\begin{algorithmic}[1]
\REQUIRE Candidate sets $\mathcal T$ and $\mathcal H$, calibration score blocks, reference outputs $O_{\mathrm{ref}}$
\ENSURE Selected parameters $(\tau^\star,h^\star)$
\STATE Initialize best objective value $\rho^\star\leftarrow +\infty$
\FOR{each $\tau\in\mathcal T$}
    \FOR{each $h\in\mathcal H$}
        \STATE Run EFQ-Softmax attention on calibration blocks with $(\tau,h)$
        \STATE Compute $\rho(\tau,h)=\|\widetilde O(\tau,h)-O_{\mathrm{ref}}\|_F/\|O_{\mathrm{ref}}\|_F$
        \IF{$\rho(\tau,h)<\rho^\star$}
            \STATE $\rho^\star\leftarrow\rho(\tau,h)$
            \STATE $(\tau^\star,h^\star)\leftarrow(\tau,h)$
        \ENDIF
    \ENDFOR
\ENDFOR
\RETURN $(\tau^\star,h^\star)$
\end{algorithmic}
\end{algorithm}

\subsection{Parameter Calibration}
\label{sec:parameter-calibration}

The affine thresholds in Eq.~\eqref{eq:d8-index} are controlled by two
global parameters, \(\tau\) and \(h\). Calibration selects these parameters
to align the threshold range with the residual distribution observed after
block-scale normalization. Specifically, \(\tau\) shifts the threshold
range, while \(h\) controls the spacing between adjacent thresholds.
Larger \(h\) produces denser thresholds and can assign more residuals to
higher E2M1 codes, whereas smaller \(h\) produces a more conservative code
distribution.

\begin{figure}[t]
    \centering
     \includegraphics[
        width=\columnwidth,
        trim=5mm 4mm 5mm 4mm,
        clip
    ]{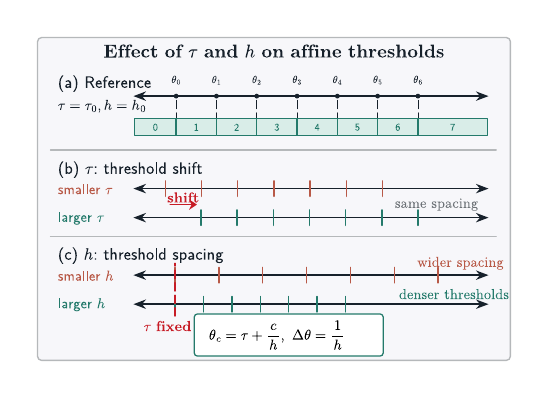}
    \caption{Effect of $\tau$ and $h$ on affine threshold placement:
    $\tau$ shifts the threshold range, whereas $h$ controls the
    spacing between adjacent thresholds.}
    \label{fig:tau-h}
\end{figure}

Fig.~\ref{fig:tau-h} illustrates how $\tau$ shifts the threshold
range and how $h$ controls the spacing between adjacent thresholds.

We select \((\tau,h)\) offline using representative attention activations.
For each candidate pair, EFQ-Softmax is applied to calibration score blocks,
and the resulting attention output is compared with the reference output:
\[
\min_{\tau,h}
\frac{\|\widetilde O(\tau,h)-O_{\mathrm{ref}}\|_F}
{\|O_{\mathrm{ref}}\|_F}.
\]
This output-level objective is used because the generated probability
operand affects both the numerator branch \(\widetilde P V\) and the
denominator branch \(\widetilde P\mathbf 1\). Scalar fitting to \(e^x\)
can be used as an initialization or diagnostic, but scalar exponential
error is not the final selection criterion.
This objective also captures the coupled effect of probability-code errors
on both the accumulated numerator and denominator, which is not reflected
by scalar exponential fitting alone.

After calibration, the selected operating point remains fixed during
inference and is validated on downstream tasks. This separation prevents
the parameter search from being interpreted as per-input adaptation or
test-set tuning.

\subsection{Algorithmic Cost and Method Boundary}
\label{sec:cost-boundary}

\paragraph{Operations removed.}
The conventional path in Eq.~\eqref{eq:post-softmax-path} first evaluates
dense elementwise exponentials for the current probability block and then
quantizes the resulting high-precision probabilities into E2M1. EFQ-Softmax
fuses these two stages into the residual-domain score-to-code rule in
Eq.~\eqref{eq:d8-index}. For each element, the EFQ-Softmax path uses
subtraction, multiplication by \(h\), floor, clipping, and bit-pattern
generation. For each microscaling block, it computes one local maximum and
one exponent-only scale. The row-level exponential used for historical
rescaling, \(\alpha_i^j=\exp(m_i^{j-1}-m_i^j)\), is outside the removed
dense elementwise path and is retained.

\paragraph{Kernel interface and placement.}
The probability-generation unit receives the shifted score block
\(x_i^j\), the microscaling partition \(\mathcal B\), and the fixed
parameters \((\tau,h)\), and returns the probability operand
\[
(x_i^j,\mathcal B,\tau,h)
\longmapsto
\left(Q_{P,i}^{j,\EtwoMone},s_{P,i}^j\right).
\]
Masks are applied before this stage when the score block
\(S_i^j=Q_iK_j^\top/\sqrt d+M_i^j\) is formed and shifted by the updated
row maximum. EFQ-Softmax is placed immediately after this shift, at the point
where the exact implementation would evaluate \(\exp(x_i^j)\). The
generated code and scale are then consumed by both the numerator branch and
the denominator branch.

\section{Experiments}
\label{sec:experiments}

\subsection{Experimental Setup}
\label{sec:exp-setup}

\paragraph{Models.}
We evaluate on three representative open models that span pure language
modeling, vision-language understanding, and text-to-video generation:
\textbf{Qwen3-8B}, \textbf{Qwen3-VL-8B-Instruct}~\cite{yang2025qwen3}, and
\textbf{WAN2.2-TI2V-5B}~\cite{wan2025wan}. For each model we replace the original
self-attention and cross-attention over the token stream with our quantized attention, leaving all
other components unchanged.

\paragraph{Quantization step.}
Our attention follows the FlashAttention-2~\cite{dao2024flashattention} tiling loop and approximates
the softmax numerator $\tilde{P}_{i}^j=\exp(S_{i}^j-m_i)$ with a block-wise
quantizer. We adopt a high-precision FP16 implementation as the baseline and
compare it against a standard MXFP4 quantization
implementation~\cite{roux2024mxfloat,zhang2024sageattention} and our
proposed EFQ-Softmax method; see the method section for the detailed
configuration of EFQ-Softmax.

\paragraph{Benchmarks and metrics.}
\emph{Qwen3-8B}: seven zero-shot classification tasks from
lm-eval-harness~\cite{eval-harness}
-BoolQ~\cite{clark2019boolq}, OpenBookQA~\cite{mihaylov2018openbookqa},
RTE~\cite{wang2019superglue}, WinoGrande~\cite{sakaguchi2021winogrande},
MMLU~\cite{hendrycks2021mmlu}, ARC-Easy and
ARC-Challenge~\cite{clark2018think}---
reported as accuracy and averaged into a 7-task mean.
\emph{Qwen3-VL-8B-Instruct}: three VQA tasks
(AI2D~\cite{kembhavi2016diagram}, OK-VQA~\cite{marino2019ok},
TextVQA~\cite{singh2019towards}, exact-match) and six referring-expression
grounding splits (RefCOCO~\cite{yu2016refcoco} and
RefCOCO$+$~\cite{mao2016generation} $\times$ \{testA, testB, val\}, ACC@0.5),
averaged into a 9-task mean.
\emph{WAN2.2-TI2V-5B}: 14 diverse text-to-video prompts at
$1280{\times}704$, 121 frames, 50 UniPC~\cite{zhao2023unipc} steps, with a fixed seed of 123469 across all prompts.

\paragraph{Parameter selection.}
For the language-model evaluation, we report two calibrated EFQ-Softmax
operating points. EFQ-Softmax-MMLU is configured with $\tau=-2.90,\,h=2.00$ and is selected for its highest MMLU accuracy in the parameter sweep, whereas EFQ-Softmax-Mean is configured with $\tau=-3.06,\,h=2.30$ and is selected for its highest seven-task mean in the parameter sweep. For Qwen3-VL, we first screen
candidate \((\tau,h)\) pairs on at most 16 samples per task and keep only
points whose per-task accuracy is within one point of MXFP4
\((\Delta\ge -0.01)\). The selected operating point,
configured with $\tau=-2.10,\,h=2.70$,
is denoted EFQ-Softmax-Balance and is used for the full vision-language evaluation.

For comparison, we also evaluate a 16-level lookup-table variant denoted EFQ-Softmax-LUT as an ablation. In contrast to our proposed 8-code configuration of EFQ-Softmax, EFQ-Softmax-LUT adopts a 16-level E2M1 code to isolate the effect of the proposed coding rule.

\subsection{Quantitative Results on Qwen3-8B}
\label{sec:exp-qwen3}

Tables~\ref{tab:qwen3-llm-1}--\ref{tab:qwen3-llm-2} report zero-shot
accuracy on Qwen3-8B at batch size 8, with the FP16  in the first row and $\Delta$ relative to the MXFP4 in parentheses. The seven tasks are split across two tables. From these results, three observations emerge.

\textbf{(i) EFQ-Softmax-LUT matches or beats MXFP4 on average.}
EFQ-Softmax-Mean reaches a 7-task mean of \(0.6773\), compared with
\(0.6749\) for the MXFP4 \((+0.0024)\). The 16-level
EFQ-Softmax-LUT ablation reaches \(0.6784\) \((+0.0036)\).
EFQ-Softmax-MMLU is on par with MXFP4 while using
the same 8-code E2M1 output set as EFQ-Softmax.
Crucially, both 8-code operating points stay
within $0.005$ of the FP16 baseline ($0.6803$), so the accuracy lost to
P-quant quantization is recovered by our simpler LUT.

\begin{table}[t]
\centering
\small
\setlength{\tabcolsep}{5.5pt}
\caption{Qwen3-8B zero-shot accuracy (FP16, batch size 8),
part~1 of 2. $\Delta$ is relative to the MXFP4 reference (row~2).}
\label{tab:qwen3-llm-1}
\begin{tabular}{l c c c c c}
\toprule
Method & BoolQ & OBQA & RTE & WinoG. & MMLU \\
\midrule
FP16     & .8657 & .3120 & .7834 & .6772 & .7300 \\
MXFP4     & .8667 & .3080 & .7726 & .6827 & .7264 \\
MXFP4-scale6    & .8648 & .3080 & .7798 & .6835 & .7255 \\
EFQ-LUT  & .8664 & .3180 & .7762 & .6851 & .7248 \\
EFQ-MMLU   & \textbf{.8703} & .3160 & .7690 & .6748 & .7223 \\
EFQ-Mean   & .8667 & \textbf{.3220} & .7762 & \textbf{.6906} & .7119 \\
\bottomrule
\end{tabular}
\begin{minipage}{\columnwidth}
\footnotesize
\vspace{2pt}
\end{minipage}
\end{table}

\begin{table}[t]
\centering
\small
\setlength{\tabcolsep}{6.5pt}
\caption{Qwen3-8B zero-shot accuracy, part~2 of 2 (ARC-Easy,
ARC-Challenge, and the 7-task mean). $\Delta$ on the mean is relative
to the MXFP4 reference.}
\label{tab:qwen3-llm-2}
\begin{tabular}{l c c c c}
\toprule
Method & ARC-E & ARC-C & Mean & $\Delta$ \\
\midrule
FP16      & .8350 & .5589 & .6803 & $+$.0055 \\
MXFP4    & .8207 & .5469 & .6749 & 0 \\
MXFP4-scale6     & .8194 & .5435 & .6749 & $+$.0001 \\
EFQ-LUT & .8325 & .5461 & \textbf{.6784} & \textbf{$+$.0036} \\
EFQ-MMLU   & .8283 & .5469 & .6754 & $+$.0005 \\
EFQ-Mean   & .8287 & .5452 & \textbf{.6773} & \textbf{$+$.0024} \\
\bottomrule
\end{tabular}
\begin{minipage}{\columnwidth}
\footnotesize
\vspace{2pt}
\end{minipage}
\end{table}

\textbf{(ii) Per-task behavior separates the two operating points.}
The two 8-code operating points exhibit different accuracy trade-offs.
EFQ-Softmax-MMLU gives the best BoolQ result in the table \((0.8703)\) and
keeps MMLU within \(0.0041\) of MXFP4. By contrast, EFQ-Softmax-Mean favors WinoGrande
\((0.6906,\,+0.0079)\) and OpenBookQA \((0.3220,\,+0.0140)\), yielding
the highest seven-task mean among the 8-code EFQ-Softmax variants.
The two $(\tau,h)$ configurations therefore deliver a tunable accuracy trade-off that MXFP4's
quantize-after-exponential scheme cannot provide.

\begin{table}[H]
\centering
\small
\setlength{\tabcolsep}{5pt}
\caption{Qwen3-8B 7-task mean across batch sizes and kernel backends.
$\Delta$ is relative to the MXFP4 reference at the same batch size.}
\label{tab:qwen3-batch}
\begin{tabular}{l c c c c}
\toprule
 & \multicolumn{2}{c}{batch = 16} & \multicolumn{2}{c}{batch = 24} \\
\cmidrule(lr){2-3}\cmidrule(lr){4-5}
Method & Mean & $\Delta$ & Mean & $\Delta$ \\
\midrule
FP16   & .6805 & $+$.0074 & .6809 & $+$.0074 \\
MXFP4  & .6731 & 0        & .6735 & 0        \\
MXFP4-scale6 & .6736 & $+$.0005 & .6736 & $+$.0001 \\
EFQ-LUT         & \textbf{.6781} & \textbf{$+$.0050} & .6757 & $+$.0023 \\
EFQ-MMLU & .6740 & $+$.0009 & .6764 & $+$.0029 \\
EFQ-Mean & .6733 & $+$.0002 & \textbf{.6785} & \textbf{$+$.0050} \\
\bottomrule
\end{tabular}
\end{table}

\textbf{(iii) Batch size does not change the
conclusion.}
Table~\ref{tab:qwen3-batch} replicates the comparison under batch sizes of 16 and 24. The seven-task average of each method deviates by no more than $0.006$ from its batch-8 result, confirming that our comparison conclusions remain stable under batching. The FP16 baseline consistently yields the highest accuracy, and the MXFP4 reference the lowest. Meanwhile, the EFQ-Softmax variants — EFQ-Softmax-LUT, EFQ-Softmax-MMLU, and EFQ-Softmax-Mean — sit in the intermediate range, with their relative ordering interchangeable within a $0.005$ accuracy band.

\subsection{Quantitative Results on Qwen3-VL-8B-Instruct}
\label{sec:exp-qwen3vl}

Tables~\ref{tab:qwen3vl-vqa}--\ref{tab:qwen3vl-grd} compare EFQ-Softmax-Balance with the MXFP4 across the full Qwen3-VL evaluation suite. The results are split into VQA (Table~\ref{tab:qwen3vl-vqa}) and referring-expression grounding (Table~\ref{tab:qwen3vl-grd}) to fit within the column width. On the three VQA tasks, the two methods yield nearly indistinguishable performance. EFQ-Softmax-Balance incurs a maximum accuracy drop of only $0.0032$ on AI2D and even achieves a slight gain of $0.0007$ on TextVQA; the worst-case degradation on VQA is therefore well below $0.01$. In contrast, EFQ-Softmax-Balance outperforms MXFP4 on all six grounding splits, with improvements ranging from $+0.0028$ (RefCOCO+ testB) to $+0.0430$ (RefCOCO testA). Aggregated over all nine tasks, the mean accuracy rises from $0.7826$ to $0.8000 (+0.0174)$. The IoU metrics underlying these ACC@0.5 scores (Table~\ref{tab:qwen3vl-iou}) follow the same trend, indicating that the quantized grounding boxes are not only correct more often but also spatially better aligned with target objects.

\begin{table}[t]
\centering
\small
\setlength{\tabcolsep}{6pt}
\caption{Qwen3-VL-8B-Instruct VQA tasks (exact match). $\Delta$ is
EFQ-Softmax-Balance minus MXFP4.}
\label{tab:qwen3vl-vqa}
\begin{tabular}{l c c c}
\toprule
Task & MXFP4 & EFQ-Softmax-bal. & $\Delta$ \\
\midrule
AI2D    & .83646 & .83323 & $-$.00324 \\
OK-VQA  & .49370 & .49184 & $-$.00186 \\
TextVQA & .79722 & .79792 & \textbf{$+$.00070} \\
\bottomrule
\end{tabular}
\end{table}

\begin{table}[t]
\centering
\small
\setlength{\tabcolsep}{5pt}
\caption{Qwen3-VL-8B-Instruct referring-expression grounding
(ACC@0.5). $\Delta$ is EFQ-Softmax-Balance minus MXFP4; every split
improves. The 9-task mean (VQA $+$ grounding) is shown in the last
row.}
\label{tab:qwen3vl-grd}
\begin{tabular}{l c c c}
\toprule
Split & MXFP4 & EFQ-Softmax-Bal. & $\Delta$ \\
\midrule
RefCOCO  testA   & .85418 & .89722 & \textbf{$+$.04304} \\
RefCOCO  testB   & .83094 & .84862 & \textbf{$+$.01768} \\
RefCOCO  val     & .85178 & .88617 & \textbf{$+$.03439} \\
RefCOCO$+$ testA & .84608 & .88051 & \textbf{$+$.03443} \\
RefCOCO$+$ testB & .74805 & .75083 & \textbf{$+$.00278} \\
RefCOCO$+$ val   & .78502 & .81367 & \textbf{$+$.02865} \\
\midrule
9-task mean      & .78260 & .80000 & \textbf{$+$.01740} \\
\bottomrule
\end{tabular}
\end{table}

\begin{table}[t]
\centering
\small
\caption{IoU of the EFQ-Softmax-Balance grounding boxes on the
RefCOCO / RefCOCO$+$ splits (Qwen3-VL full evaluation).}
\label{tab:qwen3vl-iou}
\begin{tabular}{l c}
\toprule
Split & EFQ-Softmax-bal.\ IoU \\
\midrule
RefCOCO  testA   & 0.8257 \\
RefCOCO  testB   & 0.7778 \\
RefCOCO  val     & 0.8152 \\
RefCOCO$+$ testA & 0.8058 \\
RefCOCO$+$ testB & 0.7006 \\
RefCOCO$+$ val   & 0.7527 \\
\bottomrule
\end{tabular}
\end{table}

In short, for the vision-language model, EFQ-Softmax incurs no accuracy loss on VQA and strictly outperforms MXFP4 on grounding. Moreover, it achieves a clear advantage on referring-expression splits: on these tasks, the softmax distribution over candidate boxes is sharper, which enables the method to reap benefits from the arithmetic-midpoint scale.

\subsection{Text-to-Video Generation on WAN2.2-TI2V-5B}
\label{sec:exp-wan}

\begin{figure}[h!]
    \centering
    \includegraphics[width=\linewidth]{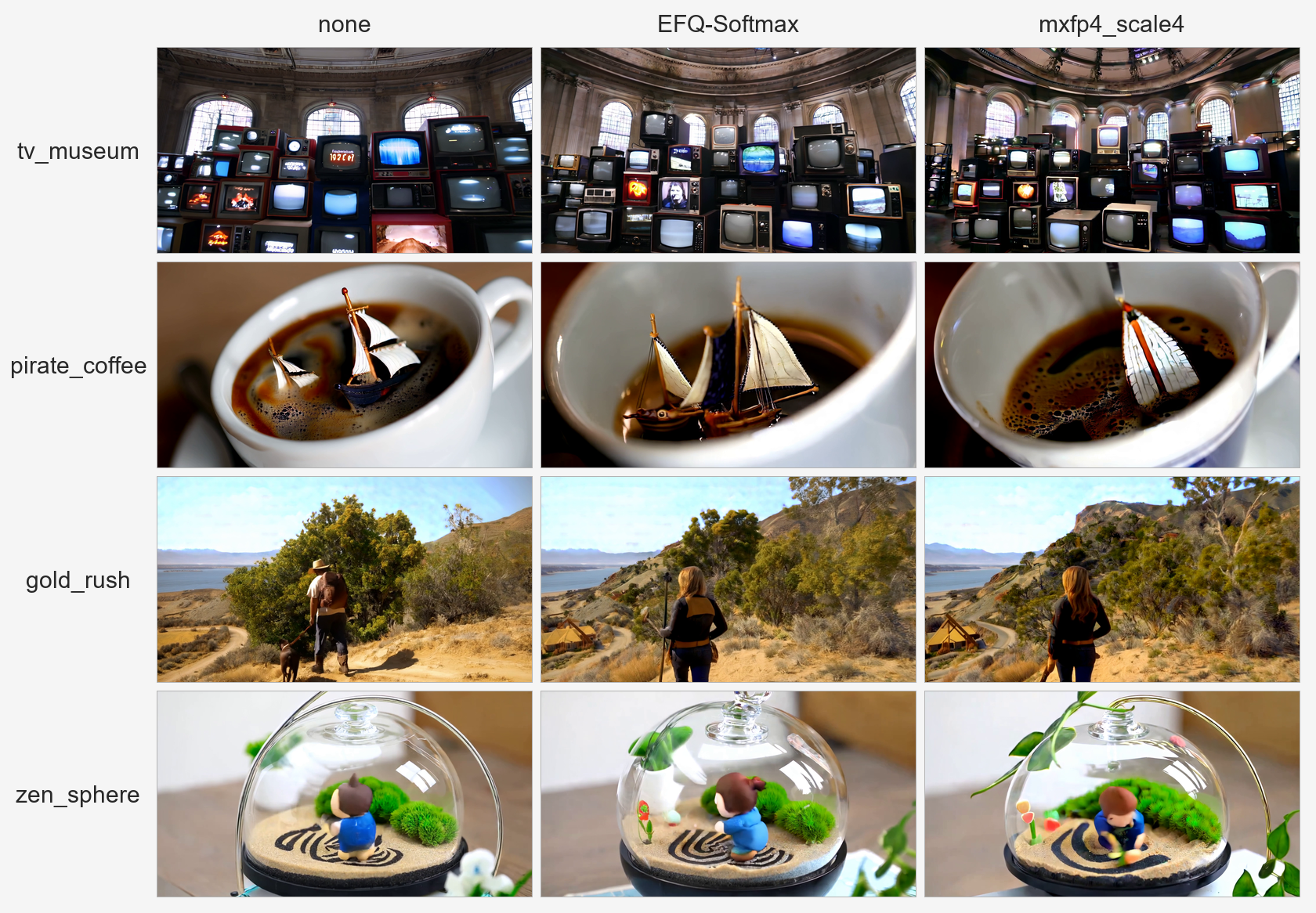}
    \caption{Middle-frame qualitative overview.}
    \label{fig:task1}
\end{figure}

\begin{figure*}[h!]
    \centering

    \begin{subfigure}[t]{0.492\textwidth}
        \centering
        \includegraphics[width=\linewidth]{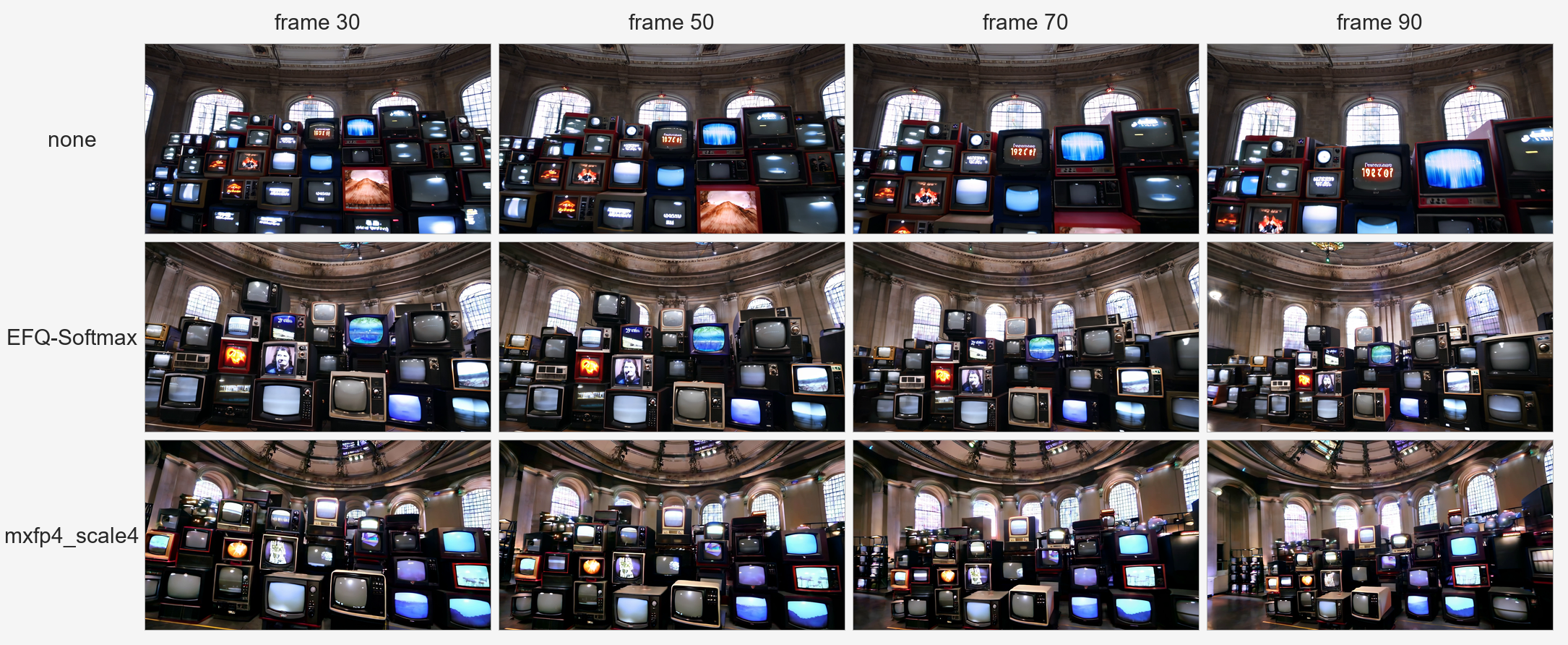}
        \caption{Television-wall scene.}
        \label{fig:temporal-tv}
    \end{subfigure}
    \hfill
    \begin{subfigure}[t]{0.492\textwidth}
        \centering
        \includegraphics[width=\linewidth]{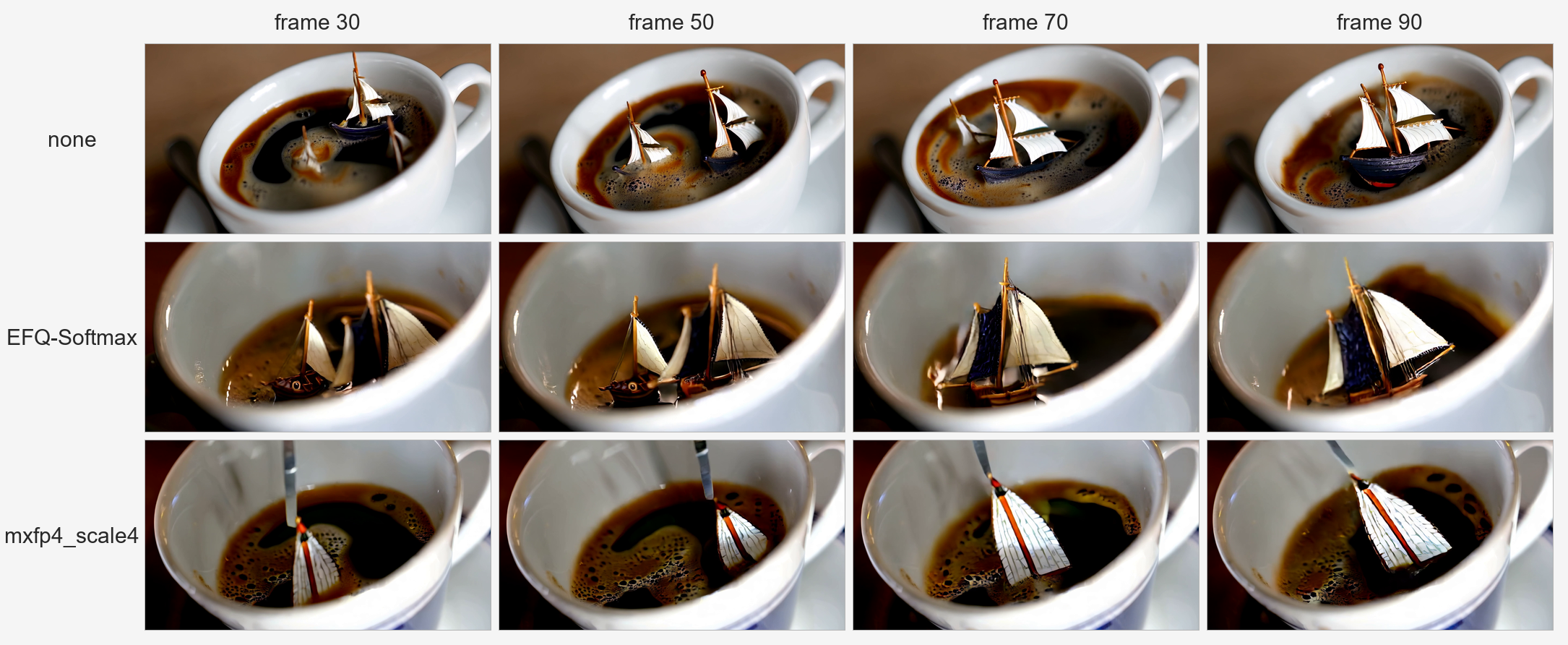}
        \caption{Coffee-cup scene.}
        \label{fig:temporal-coffee}
    \end{subfigure}

    \vspace{2mm}

    \begin{subfigure}[t]{0.492\textwidth}
        \centering
        \includegraphics[width=\linewidth]{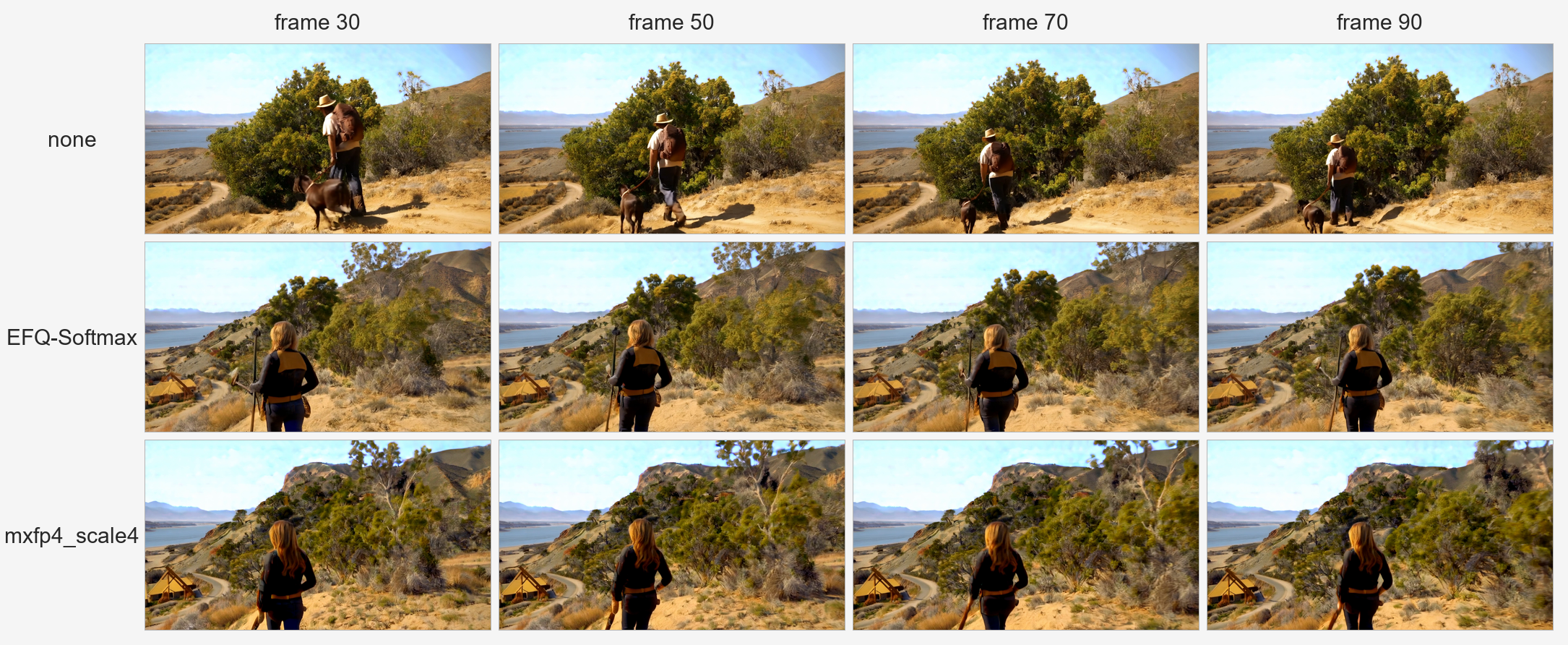}
        \caption{Outdoor scene.}
        \label{fig:temporal-outdoor}
    \end{subfigure}
    \hfill
    \begin{subfigure}[t]{0.492\textwidth}
        \centering
        \includegraphics[width=\linewidth]{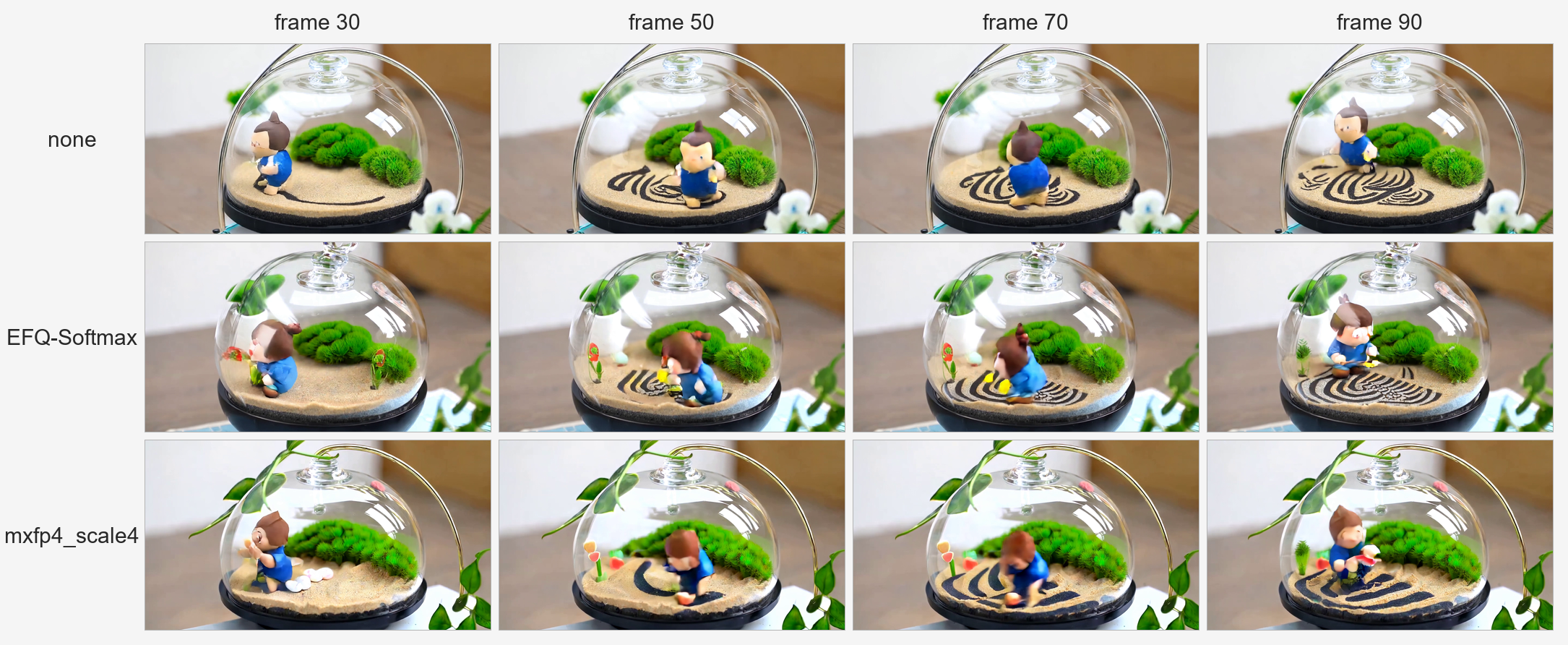}
        \caption{Terrarium scene.}
        \label{fig:temporal-terrarium}
    \end{subfigure}

    \caption{
    Temporal consistency comparison across four representative
    text-to-video examples. Columns within each panel show frames
    sampled at different timesteps, while rows correspond to different
    attention implementations.
    }
    \label{fig:temporal-consistency}
    \vspace{-2mm}
\end{figure*}
We generate 14 text-to-video clips with the WAN2.2-TI2V-5B model under three P-quant settings: the FP16 baseline, EFQ-Softmax, and \textsc{mxfp4\_scale4}. All clips adopt identical generation parameters: $1280\times 704$ resolution, 121 frames, 50 UniPC steps, and a fixed random seed of 123469. As such, the sole experimental variable is the attention kernel. The 14 prompts encompass the visual styles utilized in previous WAN2.2~\cite{wan2025wan}%
evaluations, including museum interiors, close-up object motion, archival footage, fantasy creatures, neon street scenes, drone landscapes, and papercraft/craft-material clips.

\paragraph{VBench quantitative metrics.}
To complement our qualitative comparison, we evaluate all video clips via VBench~\cite{huang2024vbench} with its default auxiliary-information configuration, which enables 10 valid dimensions covering the primary evaluation criteria. Table~\ref{tab:wan22-vbench} summarizes the results of the three variants. In terms of temporal-coherence metrics, EFQ-Softmax achieves scores comparable to both the FP16 baseline and \textsc{mxfp4\_scale4}. Subject consistency ($0.9493$), background consistency ($0.9564$), temporal flickering ($0.9784$), and motion smoothness ($0.9891$) reveal no observable degradation of temporal consistency. For visual quality dimensions, EFQ-Softmax attains the highest imaging quality ($0.7178$), surpassing the baseline ($0.7098$) and \textsc{mxfp4\_scale4} ($0.7004$), whereas its aesthetic quality ($0.6355$) lies between the two reference variants. Overall, the VBench metrics verify that EFQ-Softmax retains the visual quality of the FP16 baseline while delivering competitive performance relative to \textsc{mxfp4\_scale4}.

\begin{table}[t]
\centering
\small
\setlength{\tabcolsep}{3.5pt}
\caption{WAN2.2-TI2V-5B VBench results under the default
auxiliary-information setting (15 valid dimensions, main evaluation
criteria). \emph{Baseline} is FP16, \emph{MXFP4 scale4} is the
MXFP4 reference, and EFQ-Softmax.}
\label{tab:wan22-vbench}
\begin{tabular}{l c c c}
\toprule
VBench dimension & Baseline & MXFP4 scale4 & EFQ-Softmax \\
\midrule
subject consistency      & 0.9509 & 0.9536 & 0.9493 \\
background consistency   & 0.9559 & 0.9624 & 0.9564 \\
temporal flickering      & 0.9801 & 0.9775 & 0.9784 \\
motion smoothness        & 0.9901 & 0.9887 & 0.9891 \\
dynamic degree           & 0.5714 & 0.5000 & 0.5000 \\
aesthetic quality        & 0.6488 & 0.6295 & 0.6355 \\
imaging quality          & 0.7098 & 0.7004 & \textbf{0.7178} \\
temporal style           & 0.2645 & 0.2475 & 0.2562 \\
appearance style         & 0.2431 & 0.2416 & 0.2406 \\
overall consistency      & 0.2645 & 0.2475 & 0.2562 \\
\bottomrule
\end{tabular}
\end{table}

\paragraph{Visual quality.}
Since a single scalar metric cannot comprehensively characterize text-to-video generation quality, we present side-by-side video comparisons below. Figure~\ref{fig:task1} shows middle-frame examples for four prompts, and Figure~\ref{fig:temporal-consistency} compares the temporal consistency of different attention methods on these cases.
Each of the 14 prompts yields three video clips (FP16 baseline,
EFQ-Softmax, \textsc{mxfp4\_scale4}) generated using identical prompts and random seeds. Qualitatively, videos produced by EFQ-Softmax retain the motion coherence, lighting, and fine texture details seen in the
FP16 baseline, achieving performance comparable to \textsc{mxfp4\_scale4}. Importantly, EFQ-Softmax relies on a streamlined probability-code generation pipeline based on a single affine index rule, which differs from MXFP4’s scale-and-round procedure.

\subsection{Kernel-Level Speedup on the \mbox{Ascend~950PR} Vector Unit}
\label{sec:exp-speed}

We benchmark the fused P-quant kernel on the \mbox{Ascend~950PR} vector unit. We compare the MXFP4 with the fused EFQ-Softmax implementation across six sequence lengths spanning $16$K to $128$K tokens. The P-quant computation pipeline is split into three stages: matrix-multiply-accumulate (\emph{mmad}), fixed-point pipeline (\emph{fixpipe}), and vector processing (\emph{vector}). By construction, the \emph{mmad} stage is identical for both variants. Both implementations dequantize operands to the identical E2M1 format for matrix multiplication. Accordingly, their measured latencies align down to the microsecond, and we exclude this stage from Table~\ref{tab:speed-stage}.

\begin{table}[t]
\centering
\small
\setlength{\tabcolsep}{3.5pt}
\caption{\mbox{Ascend~950PR} vector-unit per-stage timing ($\mu$s) and per-stage speedup
for \emph{fixpipe} and \emph{vector}: MXFP4 baseline vs.\ EFQ-Softmax. Speedup is $1-t_{\mathrm{LUT}}/t_{\mathrm{base}}$.}
\label{tab:speed-stage}
\begin{tabular}{l r r r r r r}
\toprule
 & \multicolumn{3}{c}{fixpipe} & \multicolumn{3}{c}{vector} \\
\cmidrule(lr){2-4}\cmidrule(lr){5-7}
Seq & base & LUT & gain & base & LUT & gain \\
\midrule
16K  &  106.1 &  105.4 & 0.69\% &  112.6 &   67.4 & \textbf{40.10\%} \\
24K  &  243.9 &  242.1 & 0.71\% &  250.1 &  149.7 & \textbf{40.12\%} \\
32K  &  431.2 &  429.8 & 0.32\% &  443.8 &  264.4 & \textbf{40.44\%} \\
48K  &  977.2 &  969.0 & 0.84\% &  994.4 &  592.7 & \textbf{40.39\%} \\
64K  & 1738.0 & 1720.5 & 1.01\% & 1764.1 & 1050.5 & \textbf{40.45\%} \\
128K & 6955.0 & 6883.2 & 1.03\% & 7043.8 & 4192.3 & \textbf{40.48\%} \\
\midrule
mean & 1741.9 & 1725.0 & 0.77\% & 1768.1 & 1052.8 & \textbf{40.33\%} \\
\bottomrule
\end{tabular}
\end{table}

The observed speedup is consistent with the underlying LUT design. The \emph{fixpipe} stage executes per-element scaling, clipping and rounding logic, achieving a $0.3\%$--$1.0\%$ performance improvement. This modest yet steady gain arises from substituting the abs-max/log/ceil/bucketize pipeline with a single linear index lookup. The most significant improvement occurs within the \emph{vector} stage: the fused LUT reduces latency by an average of $40.33\%$, and this reduction remains nearly invariant across all tested sequence lengths ($40.10\%$ at $16$K up to $40.48\%$ at $128$K). As the end-to-end acceleration is constrained by the proportion of P-quant computation located in the vector stage, further speedups are anticipated for tensor shapes where the softmax-numerator tile accounts for a larger share compared with the matmul operation.

\section{Conclusion}
\label{sec:conclusion}

This paper introduced EFQ-Softmax, a probability-side method for low-bit attention. Instead of computing high-precision shifted-score exponentials and then quantizing the resulting probabilities, EFQ-Softmax directly generates a block-scaled E2M1 probability operand. Its core generator selects an exponent-only scale per microscaling block, maps
residual log-domain scores to 4-bit E2M1 probability codes, and feeds the same generated operand to both the numerator branch \(\widetilde P V\) and the denominator branch \(\widetilde P\mathbf 1\). EFQ-Softmax therefore changes only the probability-code generation path. It does not define a new value-side quantizer, a new attention rule, or a pruning method. The row-maximum update, historical rescaling, high-precision branch outputs, online accumulators, and final normalization remain part of the FlashAttention-style online recurrence.

Although EFQ-Softmax changes the generation of the current unnormalized
weights, the final output remains a normalized weighted sum. The online
recurrence accumulates \(\widetilde P V\) and
\(\widetilde P\mathbf 1\), and applies the final division as in standard
online softmax. Since all values in the E2M1 probability code set are
nonnegative, the current-block denominator increments are nonnegative.
Different microscaling blocks may use different scales, but their
contributions are converted to high-precision branch outputs before being
accumulated into the common online state. The only data-dependent metadata
introduced by EFQ-Softmax is the scale exponent \(k_b\) per microscaling
block; the parameters \((\tau,h)\) are fixed after calibration and are not
stored per token, per row, or per block.

Across Qwen3-8B, Qwen3-VL-8B-Instruct, and WAN2.2-TI2V-5B, EFQ-Softmax
preserves task accuracy and output quality relative to FP16 and MXFP4
baselines. These results show that direct low-bit probability-code
generation can replace post-softmax probability quantization while
maintaining the numerical behavior required by online attention.
EFQ-Softmax therefore provides a practical path toward fully low-bit
attention kernels in which probability generation, not only \(QK^\top\)
and \(PV\), is aligned with low-bit execution.

\bibliographystyle{IEEEtranS}
\bibliography{refs}

\end{document}